# A Multimodal Approach to Predict Social Media Popularity

Mayank Meghawat
Nirma Institute of Technology
Ahmedabad, India
mayank.meghawat@gmail.com

Satyendra Yadav
Delhi Technological University
Delhi, India
satyendra.dce@gmail.com

Debanjan Mahata
UALR
Arkansas, USA
dxmahata@ualr.edu

Yifang Yin
National University of Singapore
Singapore, Singapore
yifang@comp.nus.edu.sg

Rajiv Ratn Shah
IIIT-Delhi
Delhi, India
rajivratn@iiitd.ac.in

Roger Zimmermann
National University of Singapore
Singapore, Singapore
rogerz@comp.nus.edu.sg

## Abstract

*Multiple modalities represent different aspects by which information is conveyed by a data source. Modern day social media platforms are one of the primary sources of multimodal data, where users use different modes of expression by posting textual as well as multimedia content such as images and videos for sharing information. Multimodal information embedded in such posts could be useful in predicting their popularity. To the best of our knowledge, no such multimodal dataset exists for the prediction of social media photos. In this work, we propose a multimodal dataset consisting of content, context, and social information for popularity prediction. Specifically, we augment the SMP-T1 dataset for social media prediction in ACM Multimedia grand challenge 2017 with image content, titles, descriptions, and tags. Next, in this paper, we propose a multimodal approach which exploits visual features (i.e., content information), textual features (i.e., contextual information), and social features (e.g., average views and group counts) to predict popularity of social media photos in terms of view counts. Experimental results confirm that despite our multimodal approach uses the half of the training dataset from SMP-T1, it achieves comparable performance with that of state-of-the-art.*

## 1. Introduction

Huge volumes of social media content are produced by platforms such as Twitter, Facebook, Instagram, among others. People often respond to such content by viewing, liking, commenting, and sharing due to the social nature of these platforms [15, 23]. Predicting the popularity of a photo on being posted in a social media platform can find significant uses in the domains of content recommendation, advertisement, information retrieval, among others. Modeling the prediction of popularity of photos have attracted a lot of attention in research community. ACM Multimedia Conference, 2017 presented a social media prediction task (SMP-T1) in the form of a grand challenge. Several contributions [12, 28, 10, 8] were submitted in this challenge. However, most of them just focused on information provided in the SMP-T1 dataset which includes average counts for views, comments, tags, groups, members, and lengths of title and description. Since one modality might not be enough to solve a complex problem like popularity prediction, some of the submissions [6, 13] also used image content. Motivated by such work, first, we propose a new dataset, called *multimodal-SMP-dataset*, by augmenting the existing SMP-T1-dataset with additional contextual information such as titles, descriptions, and tags of social media photos in addition to crawling image content (actual photos). Next, we propose a multimodal technique for social media popularity prediction by exploiting multimodal information.

In SMP-T1 task, researchers need to provide their solutions to predict popularity of Flickr photos based on the information (first 11 fields) provided in Table 1. URLs of photos are also provided. Approximately half of the image links are broken in the original SMP-T1-dataset. Therefore, we use only 200K Flickr photos out of 432K photos in training but keep the same test set for a fair comparison with the state-of-the-art systems. Moreover, since SMP-T1-dataset (with 432K photos) does not provide much content and contextual information, we add *tags*, *title*, and *description* in addition to image content in the proposed multimodal-SMP-dataset (with 200K photos). This will help researchers who would like to leverage multimodal information in solving social media popularity prediction problem.

| Labels | Descriptions |
|---|---|
| pid | unique picture id |
| uid | user id allocated to individual user |
| postdate | date when image was posted |
| commentcount | number of comments on post |
| haspeople | 0 or 1 describing if contains people |
| titlelen | character length of the title |
| deslen | character length of the description |
| tagcount | number of tags in the post |
| avgview | user average view |
| groupcount | user is part of how many social groups |
| avgmembercount | connected with average number of members |
| *title* | title of photo |
| *tags* | tags of photo |
| *description* | description of photo |

Table 1: First 11 fields belong to SMP-T1-dataset and last three fields (in *italic* fonts) are additional fields that are added to the newly proposed multimodal dataset.

In order to predict the popularity of a social media post, we introduce two concepts. Particularly, first, we introduce a dictionary-based scoring technique for tags and title, and second, a sentiment analysis technique from description for a better understanding of the post. We try several approaches to solve this problem. First, we build a random forest regressor to understand the social information trends of a post. Next, we apply CNN model and achieve an improvement in the performance. For image content, we decide to use Transfer Learning in our solution after trying several basic models. We find that InceptionResnetV2 model produces the best results for images. Finally, we combine CNN and InceptionResnetV2 models and achieve the best results. We evaluated our method on the test set provided by the organizers as the part of *SMP-T1*. Experimental results confirm that the performance of our proposed method is comparable to the state-of-the-art inspite of using half of the training dataset.

## 2. Related Work

In recent years, several works have focused on predicting popularity from text-based [27, 7], video-based [26, 2], and photo-based [29, 11, 4] social media platforms. Most of them have employed a similar pipeline to compute popularity scores for different types of social media content (*e.g.*, tweets, photos, and videos). First, they extract relevant features, and then employ a regressor. Since in this paper we mainly focus on predicting popularity of social media photos, we present a brief review of the literature related to it. Furthermore, we also investigate some deep learning based models for popularity prediction in social media. Moreover, most of the popularity prediction studies only consider single modality (say, only text) into account without exploring other modalities (say, image content), which limits their performance of popularity prediction.

Mathioudakis and Koudas [16] detected popularity trends by identifying and grouping bursty keywords in Twitter. In order to further improve popularity prediction, Rsheed and Muhammad [19] built a model that can predict the popularity of news articles on Twitter by first classifying features of news articles into internal and external features and then use decision tree for prediction. Some work have also been proposed to predict popularity of photos. For instance, Kholsa *et al.* [11] predicted the popularity of photos by learning regression on both image content and user context. Recent studies [17, 4] confirm that the aesthetic value of photos is also very useful in predicting their popularity.

Due to immense success of Convolutional Neural Network (CNN) for challenging tasks of *classification*, *object detection*, *segmentation*, etc., it has now gained widespread popularity [3, 14]. Since not many attempts have been made for popularity prediction of photos by using CNNs, we explored their use for the continuous output regression tasks like prediction. With the advancements and improvements in the deep learning concepts and models, it has became quite easy to modify the high-end pre-trained deep learning model for a specific problem. With the introduction of *transfer learning*, we have taken the weights of pre-trained model directly and train few layers as per our requirement for desired output of the specific problem. The state-of-the-art deep learning models such as VGG16, VGG19, ResNet50, InceptionV3, Xception, InceptionResnetV2, *etc.* have shown promising results in the field of image classification [24, 9, 5]. With some modifications, we use such models in popularity prediction of social media photos.

Multimodal information has shown to be useful in a number of significant multimedia-based analytics problem [20, 21, 22]. Recent studies confirmed that it is useful in social media popularity. For instance, Wu *et al.* [31, 30, 29] have shown the importance of time information in social media popularity. This motivated us to explore multiple modalities such as information from text, photos, and videos for improving popularity prediction [18].

## 3. Our Methodology

To predict the popularity of social media posts (*e.g.*, photos), our proposed system exploits multimodal information. We try different approaches for social media popularity prediction. First, we use a random forest approach on available social and contextual information (see Section 3.1). Next, we apply a CNN model on available social and contextual information (see Section 3.2). Moreover, we apply transfer learning on image content information using the pre-trained InceptionResnet V2 model (see Section 3.3). Finally, in our multimodal approach, we combine all features derived from earlier steps and apply a convolutional model explained in Section 3.2 (see Section 3.4).

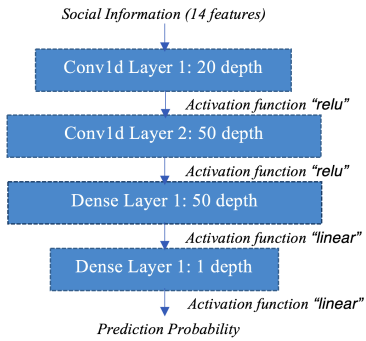

Figure 1: CNN model applied on CON-SOC Information.

### 3.1. Applying Random Forest on Contextual and Social (CON-SOC) Information

Random Forest is among the most widely used machine learning technique for predictive analytics. It is a supervised algorithm that can be used for both classification and regression. In general, it is an ensemble model, which makes predictions based on combination of base models (*i.e.*, basically a Decision Tree). We construct the base models using different subsamples of the data. In our proposed approach, we train our model on the dataset (see Section 4.1) based on mean-squared error as the evaluation criteria. Specifically, we extract features from 11 fields (see Table 1) which are provided as the part of SMP-T1 task and 3 fields (*i.e.*, title, tags, and description) that we introduce as part of the proposed new dataset. We consider such 14 features as the representation of *CON-SOC* information. We apply standard normalization method to all the features (*i.e.*, features derived from social and contextual information) prior to applying the random forest model as the range and significance of each feature is different.

### 3.2. Applying CNN model on CON-SOC Info

First we try simple dense regression model but later extend to Convolution-1D (Conv1d) model for a better performance, as CNN has shown success in solving prediction tasks in general. We find the best fit with the model shown in Figure 1. We create a 6-layer model, which contains two Conv1d layers of depth 20 and 50, *relu* as activation function, and two dropout's after each conv1d layer of 0.1 and 0.3, respectively. Finally, we add two fully-connected dense layers of depth 50 and 1 in the end. We also tried this model by adding more layers but didn't observe any significant improvement in performance.

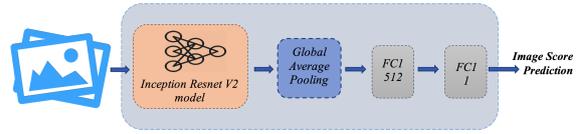

Figure 2: Modified InceptionResNet V2 model.

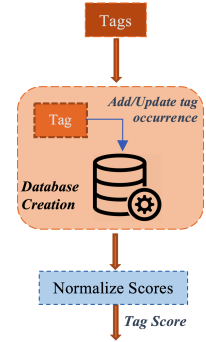

Figure 4: Tag and Title dictionary creation.

### 3.3. Applying Deep Convolutional Model on Image

Earlier studies confirm that image content is also one of the important modality to determine the popularity of photos. Thus, we can use transfer learning to derive useful information from photos based on pre-trained model on images such as InceptionResnetV2 . However, as the pre-trained models are trained for some other task, we can not directly use the same model for popularity prediction. Therefore, we had to make changes to the model, and train the same for our defined social media prediction task. The proposed model is described in Figure 2. First, we set the first 6 layer as trainable for InceptionResnetV2 model. Next, we add extra layers as per our requirement. Specifically, first, we applied the GlobalAveragePooling, next, two dense layers of depth 512 and 1, with activation function as "relu" and "linear". The reason why we set the initial 6 layers of this model as trainable is so that our model can adjust weights as per our training set images. This model is trained on only 200K images since the half of URL links are broken in the original SMP-T1 dataset.

### 3.4. Multimodal Approach

Figure 3 shows the framework of our multimodal social media popularity prediction system. It is the combination of earlier model inputs, *i.e.*, content, contextual, and social information. We include all 14 fields as described in the Section 4.1. Specifically, in addition to 11 initial features, we introduce 2 new features (tag-score and title-score) using the title and tag dictionary. Figures 4 shows the pro-cess of tag and title dictionary creation. Moreover, we add one sentiment score using a sentiment analysis tool, named, Stanford CoreNLP library [25] (see Figures 5). Figures 5 describes the process of sentiment analysis from description. We further normalize such 14 features, so that they all belong to the same scale. Furthermore, to derive useful information from the image content information, we apply the modified InceptionResNetV2 model. We pass all images in our proposed multimodal-SMP-dataset (see Section 4.1) through this model and get a prediction value as an output. We use this prediction value as the image information. We combine values from both feature scales. Now we have in total 15 features: 14 from CON-SOC information and 1 from image prediction. We pass these values to the convolutional model, which we discussed earlier (Section 3.2). The model contains 2 convolutional layers and 2 dense layers, and predict the output as final prediction. For this model, we needed to cut down the training input size from 432K to 200K, since images for most URL links were broken and since our model is using images as an input feature.

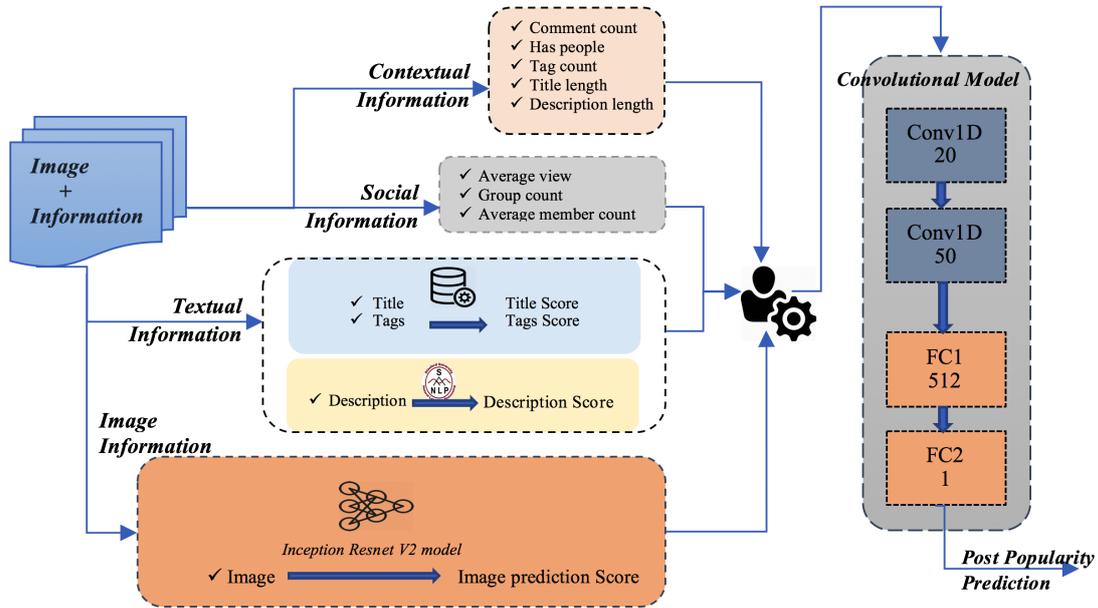

Figure 3: System framework for multimodal social media prediction.

## 4. Evaluation

### 4.1. Dataset

The official release SMP-T1 task contains entries from 135 users with around 432K posts (*i.e.*, 3.2K photos on average from a user). It includes additional information ranging to 6 years, average title length of 20, average tag count of 9, average description length 114, and average view count 131. There are total 11 fields in the original SMP-T1 task dataset (see first 11 entries in Table 1). We add 3 additional fields (*i.e.*, title, description, and tags) as contextual information, and image content to the original dataset to prepare our multimodal dataset for social media prediction task. We refer to original SMP-T1 task dataset as *SMP-T1-dataset* and our proposed dataset as *multimodal-SMP-dataset*[1] in this study. In each of the post we provide 14 fields (3 newly added features) as shown in Table 1.

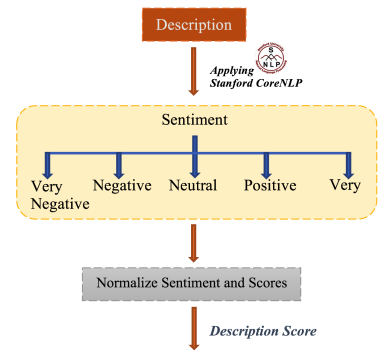

Figure 5: Sentiment analysis from description using Stanford CoreNLP library.

---

[1]It is available for research purpose at https://github.com/Macky1290/Multimodal-Social-Media-Prediction

## 4.2. Evaluation Metrics

We use Spearman RHO, Mean Absolute Error (MAE), and Mean Squared Error (MSE) evaluation metrics in our study as used in the ACM Multimedia grand challenge on SMP-T1 dataset. Spearman correlation is a nonparametric measure of the monotonicity of the relationship between two datasets. It does not assume that both datasets are normally distributed. Like other correlation coefficients, this one varies between -1 and +1 with 0 implying no correlation. Correlations of -1 or +1 imply an exact monotonic relationship. We use Scipy library to determine correlation matrix. Spearman RHO is defined by the following equation:

$$r_s = 1 - \frac{6 \sum d^2}{n(n^2 - 1)} \quad (1)$$

MAE measures the average magnitude of the errors in a set of predictions, without considering their direction. It is the average over the test sample of the absolute differences between prediction and actual observation where all individual differences have equal weight. We use Scipy library to determine correlation matrix. MAE is defined by the following equation:

$$mae = \frac{\sum_1^n |y - y_1|}{n} \quad (2)$$

MSE is a quadratic scoring rule that also measures the average magnitude of the error. It is the average of squared differences between prediction and actual observation. We use Sklearn library to determine this regression loss. MSE is defined by the following equation:

$$mse = \frac{\sum_1^n (y - y_1)^2}{n} \quad (3)$$

## 4.3. Experimental Results

We found that most of the entries of *commentcount* and *haspeople* are empty, hence, not much relevant information can be derived from them. Moreover, just considering other fields such as number of tags, length of title, and length of description, does not justify the importance of these values in determining the popularity estimate (see Table 1). This initial results motivated us to extract 3 more features (all tags, title and description) from the image path and propose a new multimodal dataset. We explored different models for social media popularity prediction. First, we performed Random Forest on normalized 14 input features. We received a spearman rho of 0.78 on validation set after applying 5-fold validation but the testing accuracy was 0.69. This model gave us an insight about the features. For instance, tag-score and title-score contributed around 10% and 18%, respectively. towards popularity prediction.

| Team | Spearman Rho | MAE | MSE |
|---|---|---|---|
| TaiwanNo.1 SMP-T1 | 0.8268 | 2.0528 | 1.0676 |
| heihei SMP-T1 | 0.8093 | 2.1767 | 1.1059 |
| NLPR_MMC_Passerby SMP-T1 | 0.7927 | 2.4973 | 1.1783 |
| BUPTMM SMP-T1 | 0.7723 | 2.4482 | 1.1733 |
| **CNN on CON-SOC** | 0.747 | 1.10 | 2.35 |
| bluesky SMP-T1 | 0.7406 | 2.7293 | 1.2475 |
| **Random Forest on CON-SOC** | 0.72 | 1.17 | 2.42 |
| WePREdictIt SMP-T1 | 0.5631 | 4.2022 | 1.6278 |
| FirstBlood SMP-T1 | 0.6456 | 6.3815 | 1.6761 |
| ride_snail_to_race SMP-T1 | -0.0405 | 9.2715 | 2.4274 |
| CERTH-ITI-MKLAB SMP-T1 | 0.3554 | 19.3593 | 3.8178 |

Table 2: Popularity prediction results on *SMP-T1-dataset*.

| Algorithm | Set | Spearman Rho | MAE | MSE |
|---|---|---|---|---|
| **Random Forest on CON-SOC Info** | Validation | 0.78 | 1.18 | 2.47 |
| | Testing | 0.69 | 1.27 | 2.98 |
| **CNN on CON-SOC Information** | Validation | 0.81 | 1.05 | 2.11 |
| | Testing | 0.73 | 1.21 | 2.51 |
| **Deep Model on image content** | Validation | 0.30 | 1.85 | 5.39 |
| | Testing | 0.27 | 1.88 | 5.59 |
| **Multimodal approach** | Validation | 0.83 | 1.00 | 2.06 |
| | Testing | 0.75 | 1.12 | 2.39 |

Table 3: Social media popularity prediction results on *multimodal-SMP-dataset*.

Next, we explored CNN on contextual and social information. We got the spearman rho to 0.81 on validation set, while 0.73 on testing set. As mentioned earlier, we tried increasing the layers and epoch, but significant gain in performance was not observed. Subsequently, we included image content in our experiment. We started with InceptionV3 and few earlier versions but got the best result with IncpetionResNetV2 model. Only using images in the mentioned model, we gained the spearman rho of 0.30 on validation and 0.27 on testing. Considering the images are not of any specific domain or background and they can belong to any field, our model performed decently and we got predictions as output.

Finally, we followed the multi-modal approach. We first trained a Deep Model on images, stored the predictions, and considered the predictions as 1 added feature to multimodal information. Thus, we have total 15 features (14 from contextual and social information and 1 from image content). We passed these 15 features to the earlier mentioned CNN model of CON-SOC information. We got Spearman's rho of 0.83 on validation set and 0.75 on testing set. This shows that combining image content and CON-SOC information results in significant performance gain [1]. Table 2 and Table 3 show the popularity prediction results on SMP-T1-dataset and Multimodal-SMP-dataset, respectively.

## 5. Conclusion

Our work represents one of the initial attempts for social media popularity prediction leveraging multimodal information. We explore different techniques on content, contextual, and social information for improving popularity prediction. First we try random forest on social and contextual information. Next, we apply CNN model on social and contextual information. Susequently, we use transfer learning using IncpetionResNetV2 on image content. Finally, we apply CNN model on content, contextual, and social information. Experimental results confirm that despite we use half of the training set, our multimodal approach give comparable performance to that of state-of-the-art. Moreover, we have provided a multimodal dataset for social media popularity prediction to research community. In future, we will work on improving our approach and expanding our multimodal dataset with additional modalities and photos.